\documentclass[sigconf,screen, nonacm]{acmart}
\renewcommand\footnotetextcopyrightpermission[1]{} 
\AtBeginDocument{%
  }

\acmConference[SIGKDD'26] {Proceedings of the ACM Knowledge Discovery and Data Mining 2026}{August 9--13, 2026}{Jeju, Korea}

\usepackage{subcaption}
\usepackage{graphicx} 
\usepackage{booktabs}
\usepackage{multirow}
\usepackage{makecell}
\usepackage[table]{xcolor}
\usepackage{colortbl}
\usepackage{xcolor}
 
\usepackage{amssymb} 
\usepackage{geometry}
\begin{document}

\title[Kirin: Improving ANN efficiency with SNN Hybridization]{Kirin: Improving ANN efficiency with SNN Hybridization}

\author{Chenyu Wang}

\affiliation{%
  \institution{Sun Yat-sen University}
  \city{Guangzhou}
  \state{Guangdong}
  \country{China}
}
\email{wangcy236@mail2.sysu.edu.cn}

\author{Zhanglu Yan}
\authornote{Corresponding authors are Zhi Zhou and Zhanglu Yan.}

\affiliation{%
  \institution{National University of Singapore}
  \country{Singapore}}
\email{zlyan@nus.edu.sg}

\author{Zhi Zhou}
\authornotemark[1]
\affiliation{%
  \institution{Sun Yat-sen University}
  \city{Guangzhou}
  \state{Guangdong}
  \country{China}
}
\email{zhouzhi9@mail.sysu.edu.cn}

\author{Xu Chen}
\affiliation{%
 \institution{Sun Yat-sen University}
  \city{Guangzhou}
  \state{Guangdong}
  \country{China}
}
\email{chenxu35@mail.sysu.edu.cn}

\author{Weng-Fai Wong}
\affiliation{%
  \institution{National University of Singapore}
  \country{Singapore}}
\email{wongwf@nus.edu.sg}

\renewcommand{\shortauthors}{Trovato et al.}

\begin{abstract}
Artificial neural networks (ANNs), particularly large language models (LLMs), demonstrate powerful inference capabilities but consume substantial energy. Conversely, spiking neural networks (SNNs) exhibit exceptional energy efficiency due to their binary and event-driven characteristics, thus motivating the study of ANN-to-SNN conversion. In this process, quantization plays a pivotal role, mapping LLMs' floating-point parameters to discrete SNN parameters via the temporal dimension of the time window. However, several challenges remain in the conversion process: (i) converting high bit-width quantization values into binary spikes requires longer time windows, increasing system latency; and (ii) the inherent trade-off between the information loss of single-spike schemes and the energy costs of multi-spike ones in SNN. To address these challenges, we propose Kirin, a integer and spike hybrid based SNN to achieve accuracy lossless ANN-to-SNN conversion with time and energy efficiency. 
Specifically, we first propose a Spike Matrix Hybridization strategy that encoding low bit-width parameters that leading to small time window size into binary spikes while preserving the rest in integer format, thereby reducing the overall latency of SNN execution. 
Second, we introduce a silence threshold mechanism to regulate the timing of single-spike firing, ensuring the output is mathematically equivalent to the LLM's output and preserves accuracy. Experimental results demonstrate that Kirin, under a W4A4\&8 quantization setting, achieves near-FP16 accuracy while reducing energy consumption by up to 84.66\% and shortening time steps by 93.75\%.
\end{abstract}



\maketitle

\section{Introduction}
Artificial Neural Networks (ANNs) represented by Large Language Models (LLMs) currently stand at the forefront of modern AI. These models have demonstrated exceptional proficiency in extracting insights from massive data repositories, revolutionizing fields such as natural language processing \cite{achiam2023gpt}, code generation \cite{joel2024survey}, and medical diagnosis \cite{xie2024llamafoundationlargelanguage}. However, the emergent capabilities arising from billions of parameters come at the cost of substantial inference energy consumption, which scales with the processing of massive input data. For instance, a 10-second ChatGPT interaction consumes approximately 10 kJ of energy (operating at ~1000 watts) \cite{de2023growing}, vastly exceeding the human brain's consumption of <200 J (<20 watts) for similar cognitive tasks \cite{roy2019towards}. Based on this observation, brain-inspired Spiking Neural Networks (SNNs) \cite{schuman2022opportunities} have emerged to bridge this energy gap. SNNs leverage binary, event-driven representations and the Integrate-and-Fire (IF) mechanism to replace intensive Multiply-Accumulate (MAC) operations with energy-efficient temporal Accumulate (ACC) operations, while simultaneously minimizing data movement overhead through sparse communication, thus offering a viable path for sustainable large-scale inference \cite{yan2026matterhornefficientanalogsparse}.

\begin{figure*}[t]
  \centering
  \begin{subfigure}[t]{0.49\textwidth}
    \centering
    \includegraphics[width=\linewidth]{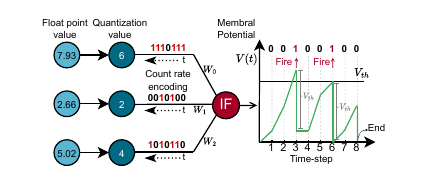}
    \caption{Rate encoding and Integrate-and-Fire mechanism. }
    \label{fig1:left}
  \end{subfigure}
  \hfill
  \begin{subfigure}[t]{0.47\textwidth}
    \centering
    \includegraphics[width=\linewidth]{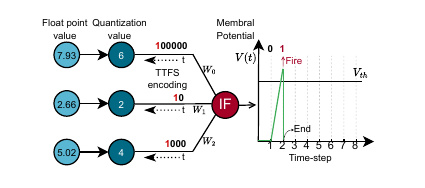}
    \caption{Time-to-First-Spike encoding and Integrate-and-Fire mechanism.}
    \label{fig1:right}
  \end{subfigure}
  \caption{From Floating-Point Values to Spike Encoding and Integrate-and-Fire mechanism. Ten floating-point values are randomly sampled and quantized (scale = 1.42, zero-point = 0); three examples are shown with their binary codes and corresponding spike trains.}
  \label{pic-encoding}
\end{figure*}

Quantization-based ANN-to-SNN conversion typically involves two stages: parameter encoding and the Integrate-and-Fire (IF) process. In the encoding phase, floating-point parameters are quantized using mixed precision to mitigate information loss: outliers in parameters are assigned high bit-widths, while normal parameters retain low bit-widths \cite{czako2025addressing, MLSYS2024_5edb57c0, MLSYS2024_42a452cb}. These parameters are then encoded into binary spikes over $T=2^{bit}$ time steps, commonly utilizing temporal coding schemes such as Time-to-First-Spike (TTFS) \cite{yan2025otters} or rate-based coding \cite{tang2024sorbet}. Subsequently, the IF neurons process the resulting spike trains and synaptic weights through temporal potential accumulation and threshold-based firing. 
However, this mechanism creates a critical bottleneck: achieving high information fidelity via quantization and encoding will lead to excessive latency and energy consumption. For instance, 8-bit quantization necessitates an extended time window ($T=256$), introducing substantial latency as the system must accommodate the maximum potential spiking time. 
Furthermore, existing coding schemes face inherent limitations in balancing precision and cost: while TTFS is energy-efficient but prone to information loss due to its over sparsity problem, maintains high accuracy but triggers heavy energy consumption due to dense spike trains. Consequently, navigating the trade-off between representation accuracy and energy efficiency becomes essential.

Current efforts to address the latency challenges of long time windows largely rely on lowering quantization precision or pruning information. For example, QFFS \cite{li2022quantization} aggressively compresses activations to 2-bit to reduce time steps, while Wu et al. \cite{wu2025optimizing} apply a dynamic cutoff during inference to drop later spikes carrying less information. Nevertheless, such strategies will sacrifice precision, leading to a degradation in model performance. Moreover, balancing the trade-off between the information loss of TTFS and the energy inefficiency of rate-encoding presents a persistent challenge. 

To address these challenges, we propose Kirin\footnote{We named our methods Kirin (or Qilin), which blending a hybridization of the lion and dragon.}, the first framework that leverages integer spike hybridization to simultaneously optimize the time and energy efficiency of ANN-to-SNN accuracy-lossless conversion. In Kirin, we formulate the matrix operations within LLMs, including Attention and MLP modules, as multiplications between a spike train matrix and a weight matrix. To address the latency caused by a small number of high bit-width values, we decompose matrix multiplications into a hybrid strategy combining dominant SNN accumulations with a small fraction of integer multiplications. Specifically, our proposed spike matrix hybridization strategy selectively retains quantization outliers that would necessitate long time windows are retained as integers and are not converted into spike trains. To minimize the overhead of integer MAC operations, we select the matrix with fewer quantized outliers necessitating integer retention as the spike matrix. Furthermore, to achieve an optimal trade-off between accuracy and energy efficiency, we introduce a Silence Threshold mechanism based on the energy-efficient TTFS, inspired by Rate encoding. We employ TTFS for temporal encoding and define the threshold in the IF phase as a function of the quantization scale, thereby mathematically ensuring equivalence to de-quantization. When the threshold is reached, the TTFS fire time increments while keeping the output silent by not firing an actual spike, continuing until no new potential accumulates. This process retains the energy efficiency of TTFS while incorporating the high inter-layer information fidelity of Rate encoding. 

Our contributions are summarized as follows:

\begin{itemize}

    \item We propose Kirin, the first integer-spike hybridization framework for spiking LLMs, designed to simultaneously optimize time and energy efficiency. 

   \item Through the Spike Matrix Hybridization strategy and Silence Threshold mechanism, we resolve the long time window issue induced by high-bit quantization with minimal energy overhead and achieve lossless accuracy with low energy cost. 


    \item Extensive experiments demonstrate that the proposed method achieves lossless conversion accuracy, outperforming state-of-the-art SNNs and quantized ANNs in both energy and time efficiency. Specifically, it attains near FP16 accuracy while reducing energy costs by 84.66\% and the time window by 93.75\%.
\end{itemize}




\section{Preliminary}
\label{Preliminary}

\subsection{ANN-to-SNN parameters mapping and encoding}

ANN-to-SNN conversion aims to transform a pre-trained ANN into a functionally equivalent SNN while preserving inference accuracy and improving energy efficiency. In this process, we adopt a quantize-then-encode paradigm, where ANN parameters are first quantized into integers and subsequently mapped to spike-based encodings, as shown in Fig. \ref{pic-encoding}. By discretizing continuous-valued ANN parameters into low-bit integers, quantization establishes a fundamental bridge to discrete spiking mechanisms. Currently, two integers-to-spike train encoding paradigms are widely used in SNNs \cite{auge2021survey}: rate-based encoding, represented by Rate Encoding \cite{diehl2015fast}, and temporal encoding, represented by TTFS Encoding \cite{mostafa2017supervised}. 

\textbf{Rate Encoding.} 
In rate-based encoding, neural weights or activations are represented by the firing frequency of a spiking neuron within a fixed time window. A commonly used rate-encoding scheme is Rate, in which the quantized value determines the total number of spikes. Specifically, given an n-bit quantized value $q$, the corresponding neuron $i$ generates a spike train whose spike count $N_i$ within a time window of length $T$ (typically $T = 2^n$) satisfies $q = N_i$. The spike rate of Rate Encoding in a time window is:
\begin{equation}
    r = \frac{N_i}{T},
\end{equation}

\textbf{Time-to-First-Spike Encoding.}
Temporal coding encodes information in the precise timing of spikes rather than their frequency. The most basic temporal coding scheme is {\em time-to-first-spike} (TTFS) encoding. In TTFS encoding, each neuron emits at most one spike, and the quantized relative time of the spike $q$ is mapped to the spike latency. In the original TTFS paradigm, larger values trigger spikes earlier (i.e., $t = T - q$). This mechanism mimics biological neurons, in which stronger stimuli lead to faster accumulation of the membrane potential and earlier firing. However, in the context of ANN-to-SNN conversion, parameter magnitude is equally critical. To make subsequent parameter encoding more intuitive and computationally efficient, we define the spike time as $t = q$. The spike rate of TTFS encoding is defined as
\begin{equation}
r = \frac{1}{T},
\end{equation}
since each neuron fires at most one spike within the time window.


\subsection{SNN Integrate-and-Fire paradigm}
Following the transformation of ANN parameters into the binary spike representations required by SNNs, the subsequent inference process adheres to the SNN computational paradigm. Among the diverse spectrum of spiking neuron models, the {\em integrate-and-fire} (IF) mechanism is widely regarded as a more robust and efficient choice due to its simplified dynamics and reduced computational overhead \cite{burkitt2006review}. The operational semantics of the IF neuron exhibit distinct characteristics when coupled with Rate Encoding and TTFS encoding. 

\textbf{Count Rate coded IF neuron model. }
Under the Rate encoding paradigm, information is encoded via spike frequency over a time window $T$. The IF neuron integrates weighted spikes into its membrane potential, firing a spike whenever a threshold $V_{\text{th}}$ is exceeded. This IF cycle repeats until the end of the time window. The discrete-time dynamics of this process can be formulated as follows:
\begin{equation}
I(t) = \sum_{i=1}^{n} W_i I_i(t),
\end{equation}
\begin{equation}
V(t) = V(t-1)+ I(t) - s(t) \cdot V_{\mathrm{th}},
\end{equation}
\begin{equation}
s(t) =
\begin{cases}
0, & \text{if } V(t-1)+ I(t) < V_{\mathrm{th}},\cr
1, & \text{if } V(t-1)+ I(t) \ge V_{\mathrm{th}}.
\end{cases}
\end{equation}
where $V(t)$ denotes the accumulated membrane potential at time step $t$, while $W_i$ and $I_i(t)$ represent the synaptic weight and the input spike from neuron $i$ at time $t$, respectively. In this configuration, the output of the SNN maintains the Rate encoding format, as illustrated in Figure \ref{fig1:left}.

\textbf{TTFS coded IF neuron model.}
For TTFS encoding, the  IF neuron model restricts the neuron to emit only a single spike, using first-spike latency as the sole code and maintaining neuronal silence thereafter . This process formalized as:

\begin{equation}
s(t) =
\begin{cases}
1, & V(t-1)+I(t) \ge V_{\mathrm{th}} \ \land\ \sum_{\tau < t} s(\tau)=0,\cr
0, & \text{otherwise}.
\end{cases}
\end{equation}
\begin{equation}
V(t)=
\begin{cases}
V(t-1)+I(t), & \sum_{\tau \le t} s(\tau)=0,\cr
0, & \sum_{\tau \le t} s(\tau)=1.
\end{cases}
\end{equation}

Under standard TTFS semantics, IF neurons are strictly constrained to emit at most a single spike per time window. Consequently, any subsequent threshold crossings driven by accumulated membrane potential are ignored, as illustrated in Fig.~\ref{fig1:right}.


\section{ANN-to-SNN Conversion Workflow}

\begin{figure*}[t] 
    \centering
    \includegraphics[width=\textwidth]{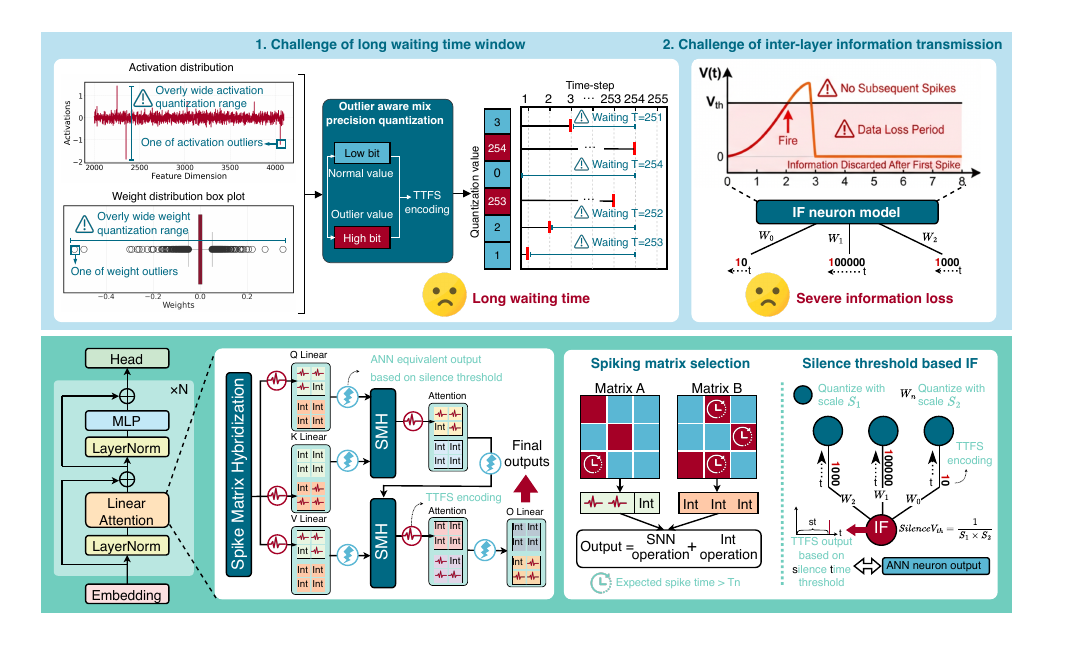} 
    \caption{Overview of the challenges and proposed ANN-to-SNN lossless conversion framework. Top: The two main bottlenecks in current SNN conversion—latency due to activation outliers and data loss in IF neurons. Bottom: The proposed solution utilizing Spike Matrix Hybridization to handle long latency and a Silence Threshold mechanism to ensure lossless outputs.}
    \label{fig:workflow}
\end{figure*}

In this section, we first analyze the challenges in the ANN-to-SNN conversion process (Section \ref{Challenges and Motivation}), focusing on parameter encoding and the IF mechanism, and then propose corresponding solutions to each (Section \ref{Spike Matrix Hybridization} and \ref{Silence Threshold based TTFS}). Finally, we presents the overall framework of the proposed Kirin (Section \ref{Kirin Overall Workflow}).

\subsection{Challenges and Motivation}
\label{Challenges and Motivation}

\textbf{Challenge 1).} During parameter encoding, a typical ANN-to-SNN conversion pipeline quantizes ANN parameters and then maps them to binary encoding (see Section \ref{Preliminary}). However, Outliers in weights and activations inflate the quantization range, thereby compressing the effective resolution for the many normal values with similar magnitudes, which degrades accuracy. A common mitigation is to apply high-bit quantization to outliers (typically 8 bit) and low-bit quantization to normal values (typically 4 bit) \cite{MLSYS2024_5edb57c0, MLSYS2024_42a452cb}. In ANN-to-SNN conversion setting, however, the time-window length serves as an information carrier equivalent to the quantization bit-width and scales exponentially with it ($T = 2^{bit}$). Consequently, within a group of neurons, outliers quantized at higher bit-widths, when mapped to binary temporal codes (e.g., TTFS), become the bottleneck for the waiting time of the corresponding IF neuron model, substantially increasing inference latency and reducing overall throughput.

\textbf{Challenge 2).} In IF neuron model, TTFS imposes a structural constraint that permits at most one spike within a time window. An IF neuron fires upon the first threshold crossing and remains inactivate for the remainder of the window, even if subsequent weighted membrane-potential accumulation could reach the threshold, it cannot be encoded as additional spikes. This induces severe information truncation across layers, rendering the layer-wise representations mathematically non-closed, which leads to inference failure or substantial performance degradation. Although Rate Count Encoding does not suffer from the aforementioned issue, its encoding scheme yields spike rates that are several times higher than those of TTFS, resulting in multiplicatively increased energy consumption. This outcome runs counter to the original motivation of ANN-to-SNN conversion, which is to reduce energy usage.

In summary, a lossless conversion from ANN to time and energy efficient SNNs requires tackling: (i) outlier-driven expansion of the quantization dynamic range and the ensuing heterogeneous bit-width quantization that inflates time windows and creates system-level latency bottlenecks, and (ii) the trade-off between TTFS-induced cross-layer information loss from the single-spike constraint and the elevated energy cost of higher spike rates, for example in Rate Count Encoding.

\subsection{Spike Matrix Hybridization}
\label{Spike Matrix Hybridization}

In the parameter encoding phase, to minimize information loss induced by quantization, we first employ the MAD algorithm \cite{823d1d7b-04be-3f06-8541-91bd1cd4e33f} —owing to its robustness against extreme values— offline to detect outliers within both weights and activations, subsequently adopting an outlier-aware hybrid precision quantization. Although TTFS introduces limitations at the IF inference stage, we nevertheless choose TTFS for parameter encoding, as its inherently low spike rate yields additional energy savings for the SNN. After the quantized parameters are converted into TTFS encoding, the maximum time windows for normal values and outlier values are defined as:
\begin{equation}
T_n=2^{b_n}, \quad T_o=2^{b_o},
\end{equation}
where $b_n$ denotes the low quantization bits for normal values, and $b_o$ denotes the high quantization bits for outlier values. Within a single IF neuron model, the latency bottleneck is typically determined by $T_o$, as illustrated in Fig. 2. Consequently, our work focuses on shortening $T_o$ while minimizing losses including accuracy degradation and energy cost.

A straightforward approach is to bypass TTFS encoding for outliers with excessively long time windows, and instead use their quantized integer values directly for matrix multiplication. A potential concern is whether retaining a significant number of MAC operations might undermine the energy-efficiency advantages of SNNs. However, our empirical analysis alleviates this concern, revealing that outliers constitute only a negligible fraction of the total parameters (e.g., merely 2.68\% in Llama2-7B). Nevertheless, we remain committed to minimizing the quantity of retained integer values. In the linear attention and MLP layers, all matrix transformations can be viewed as processes where multiple spike train neurons perform IF operations with their corresponding synaptic weights. In this context, selecting a proper matrix to transform as the spike train neuron is critical for shortening $T_o$.

Let the two matrices be $A$ and $B$. For any matrix $M \in {A, B}$, and its elements be denoted as $m_{i,j}$. For any outliers whose converted TTFS spike time would exceed the maximum time step $T_n$ of normal values, we bypass the encoding process and retain the element in its quantized integer format. Consequently, the number of such outliers in each matrix can be expressed as:

\begin{equation}
C(M) = \sum_{i,j} \mathbb{I}\left((i,j)\in \mathcal{O}_M \land |m_{i,j}|>  T_n\right),
\end{equation}
\begin{equation}
\mathcal{O}_M=\big\{(i,j)\mid m\{i,j\}\text{ is an outlier}\big\}
\end{equation}

By comparing $C(A)$ and $C(B)$, we select the matrix with the smaller $C$ value as the spike neuron (while retaining a portion of quantized integers), which implies the minimum number of integer multiplications required. The final matrix multiplication $Y = A \times B$ can be reconstructed as the sum of the following two components:

\begin{equation}
Y = \underbrace{\bigl(IF(A^{(S)} , B)\bigr)}_{\text{SNN Operation}}+
\underbrace{\bigl(A^{(I)} \times B\bigr)}_{\text{INT Operation}}.
\end{equation}

Our experiments demonstrate (see Appendix C) that only a very small fraction of values in the matrix remain as integers, ranging from 1.61\% to 2.14\%.

\subsection{Silence Threshold based TTFS}
\label{Silence Threshold based TTFS}

Upon the determination of the spike train matrix, the computation proceeds to the IF neuron model stage. In the context of TTFS encoding, the integration phase necessitates waiting for all spiking neurons within the spike matrix to fire, subsequently performing a weighted accumulation onto the membrane potential. Assuming there are $n$ spiking neurons, this process can be formulated as follows:
\begin{equation}
\label{eq12}
V(t)=V(t-1)+\sum_{i=1}^{n} a_i(t)W_it_i
\end{equation}
\begin{equation}
a_i(t)=\begin{cases}
1 &\text{if neuron $i$ spikes at $t$}\cr
0 &\text{else}
\end{cases}
\end{equation}
where $V(t)$ denotes the accumulated membrane potential at time step $t$, $W_i$ represents the quantized synaptic weight corresponding to spiking neuron $i$, and $t_i$ indicates the timestamp at which neuron $i$ emits a spike. According to the TTFS encoding rules, $t_i$ also signifies the quantized value represented by this spike. This potential accumulation process persists until either all spiking neurons have fired or the time window is exhausted. Through the processing of Spike Matrix Hybridization, the maximum duration of potential accumulation is bounded by $T_n$.
Regarding the integer parameter components within the spike matrix (assuming $m$ components), a matrix multiplication is performed directly with their corresponding quantized weights. The resulting products are then directly added, in the form of potential, to the final accumulated membrane potential:
\begin{equation}
\label{eq14}
V_{total}=V(T)+\sum_{j=1}^{m} W_jI_j
\end{equation}
where $V(T)$ is the total potential accumulated from the spiking neurons, while $I_j$ and $W_j$ represent the neuron value and the corresponding weight of the integer component, respectively.

In the firing phase, to achieve an output equivalent to that of an ANN and ensure lossless accuracy, we draw inspiration from SpikeQuant \cite{wang2025energy} and set the firing threshold $V_{th}$ to $\frac{1}{S_1 \cdot S_2}$. Here, $S_1$ and $S_2$ denote the quantization scales for Matrix A (input neurons, representing either spike trains or integer values) and Matrix B (synaptic weights), respectively. Unlike SpikeQuant, which utilizes Rate encoding for its output, our approach maintains the output of the IF neuron in the more energy-efficient TTFS coding format.

In an ANN, the final output of two quantized matrices A and B of length $g$, following matrix multiplication and dequantization, is expressed as:
\begin{equation}
\label{eq15}
Output_{ANN}=\underbrace{S_1\cdot S_2}_{z_1}\underbrace{\sum_{i=1}^{g} a_i\cdot b_i}_{z_2}+bias
\end{equation}
where $a_i$ and $b_i$ are the quantized values within Matrix A and B. Deriving from Equations (\ref{eq12}) and (\ref{eq14}), the accumulated potential $V{total}$ resulting from the hybrid execution—combining dominant SNN accumulations with sparse integer multiplications—is identical to the $z_2$ component in Equation (\ref{eq15}). In SNNs, the firing result is mathematically equivalent to $\left \lfloor \frac{V_{total}}{V_{th}} \right \rfloor$. By setting $V_{th}=\frac{1}{S_1 \cdot S_2}$, the number of times the threshold is reached equals the integer part of $Output_{ANN}$. The residual potential $V_{rest}$ that does not reach $V_{th}$, when divided by $V_{th}$, corresponds to the fractional part of $Output_{ANN}$.

In traditional TTFS, the IF process terminates immediately after the potential reaches $V_{th}$ and triggers a spike. Conversely, in Rate encoding, a spike is triggered every time $V_{th}$ is reached until no further potential accumulates. We integrate the firing characteristics of both by introducing silence time ($st$) and a silence threshold ($S_{th}$). Numerically, $S_{th}$ is equal to $V_{th}$. The distinction lies in the behavior: when the membrane potential reaches $S_{th}$, the neuron does not fire immediately; instead, $st$ is incremented by 1. This process continues until all spiking neurons have emitted spikes or the maximum time window $T_n$ is reached. Only then does the IF neuron model determine the specific time step to fire based on the value of $st$. Consequently, the final output of the IF neuron model is:
\begin{equation}
Output_{IF}=st+\frac{V_{rest}}{S_{th}}+bias=Output_{ANN}
\end{equation}
This guarantees that the SNN achieves the same output as the ANN while utilizing the energy-efficient TTFS coding scheme, thereby ensuring lossless accuracy.

\subsection{Kirin Overall Workflow}
\label{Kirin Overall Workflow}
As illustrated in the bottom panel of Figure \ref{fig:workflow}, within the linear attention module, both the linear projection and attention computation phases can be formulated as matrix multiplications between two matrices. In the linear projection phase, the input activation matrix and the weight matrix are first processed by the Spike Matrix Hybridization module. This module dynamically selects the matrix containing fewer quantized values that would necessitate a long inference time window as the spike matrix. Subsequently, quantized values within this spike matrix that are prone to inducing long latencies are retained as integers for subsequent computation, while the remaining values are encoded using TTFS. This hybrid strategy aims to mitigate the long time window challenge inherent in ANN-to-SNN conversion while minimizing energy costs.

Following matrix selection, the spike matrix and weight matrix are processed via the SNN's IF neuron model. Here, a Silence Threshold mechanism is employed to ensure that the TTFS-based SNN output is mathematically equivalent to the floating-point output obtained from the dequantized ANN matrix. Thereafter, one set of SNN outputs (e.g., Q linear projection) and its associated set (e.g., K linear projection) are fed back into the  Spike Matrix Hybridization module to redetermine the spike and weight matrices. This process iterates until the final output is generated.

Although the bottom panel of Figure \ref{fig:workflow} depicts the workflow of Kirin within the linear attention module of LLMs, the method is equally applicable to the MLP components of LLMs, as Kirin fundamentally optimizes the matrix multiplication operations in ANN-to-SNN conversion.

\begin{table*}[t]
  \centering
  \caption{Perplexity ($\downarrow$) comparison of Llama2 and OPT models using different quantization and SNN methods.}
  \label{tab:main_ppl_comparison}
  
  \begin{subtable}{\linewidth}
      \centering
      \small 
      \setlength{\tabcolsep}{3.5pt} 
      \caption{Perplexity of Llama2 Models}
      \label{tab:llama_ppl}
      \begin{tabular}{ll ccc l cccc}
        \toprule
        \multirow{2}{*}{Methods} & \multirow{2}{*}{Bits} & Weight & Activation & SNN & \multirow{2}{*}{Timestep} & \multicolumn{2}{c}{Llama2-7B} & \multicolumn{2}{c}{Llama2-13B} \\
        & & outlier & outlier & conversion &  & WikiText & C4 & WikiText & C4 \\
        \midrule
        Fp16 & W16A16 & $\times$ & $\times$ & $\times$ & - & 5.68 & 7.08 & 4.88 & 6.46 \\
        \midrule
        SmoothQuant & W4A4 & $\times$ & $\checkmark$ & $\times$ & - & 22.62 & 31.21 & 33.98 & 41.53 \\
        OmniQuant & W4A4 & $\times$ & $\times$ & $\times$ & - & 14.61 & 19.35 & 12.3 & 15.87 \\
        Atom & W4A4 & $\times$ & $\checkmark$ & $\times$ & - & 6.16 & 7.70 & 5.46 & 7.03 \\
        SpikeLLM & W4A(4\&8) & $\times$ & $\checkmark$ & $\checkmark$ & T=256 & 11.36 & 15.87 & 9.71 & 12.10 \\
        SpikeQuant & W4A(4\&5) & $\times$ & $\checkmark$ & $\checkmark$ & T=32 & 5.80 & 7.33 & 5.04 & 6.65 \\
        \rowcolor{gray!10}
        Kirin & W4A(4\&8) & $\checkmark$ & $\checkmark$ & $\checkmark$ & T=16 & \textbf{5.71} & \textbf{7.25} & \textbf{5.03} & \textbf{6.61} \\
        \bottomrule
      \end{tabular}
  \end{subtable}
  
 \vspace{1em} 
  \begin{subtable}{\linewidth}
      \centering
      \small 
      \setlength{\tabcolsep}{3.5pt} 
      \caption{Perplexity of OPT Models}
      \label{tab:opt_ppl}
      \begin{tabular}{ll ccc l cccc}
        \toprule
        \multirow{2}{*}{Methods} & \multirow{2}{*}{Bits} & Weight & Activation & SNN & \multirow{2}{*}{Timestep} & \multicolumn{2}{c}{Opt1.3B} & \multicolumn{2}{c}{Opt2.7B} \\
        & & outlier & outlier & conversion & & WikiText & C4 & WikiText & C4 \\
        \midrule
        Fp16 & W16A16 & $\times$ & $\times$ & $\times$ & - & 14.63 & 14.72 & 12.47 & 13.17 \\
        \midrule
        OneBit & W1A16 & $\times$ & $\times$ & $\times$ & - & 25.42 & 22.95 & 21.86 & 20.76 \\
        LRQuant & W2A1 & $\times$ & $\times$ & $\times$ & - & 48.43 & 69.34 & 30.59 & 42.49 \\
        GPTQ & W3A16 & $\times$ & $\times$ & $\times$ & - & 20.97 & 21.63 & 16.88 & 18.17 \\
        Vanilla Quantization & W4A4 & $\times$ & $\times$ & $\times$ & - & 35.61 & 30.52 & 15.90 & 16.18 \\
        SpikeQuant & W4A(4\&5) & $\times$ & $\checkmark$ & $\checkmark$ & T=32 & 16.38 & 16.56 & 13.15 & 13.91 \\
        \rowcolor{gray!10}
        Kirin & W4A(4\&8) & $\checkmark$ & $\checkmark$ & $\checkmark$ & T=16 & \textbf{16.33} & \textbf{16.26} & \textbf{12.89} & \textbf{13.72} \\
        \bottomrule
      \end{tabular}
  \end{subtable}
\end{table*}

\section{Experiments}

\subsection{Experimental setup}
\textbf{Training Details.} In the ANN-to-SNN conversion pipeline, quantization of both activations and weights is a prerequisite. We employ a mixed-precision quantization strategy where weights and activations are primarily quantized to 4-bit, while outliers are processed at 8-bit precision. All quantization and conversion experiments were conducted on two NVIDIA A100 40GB GPUs. To evaluate inference energy efficiency, we utilize energy measurements for computation, data movement, and memory access, derived from a commercial 22 nm process.

\textbf{Evaluation Baselines.} Given the limited number of existing quantization-based ANN-to-SNN conversion methods, we benchmark our approach against two state-of-the-art SNN conversion methods: SpikeLLM \cite{xing2024spikellm} and SpikeQuant \cite{wang2025energy}, as well as seven established ANN quantization techniques across various bit-widths. Specifically, for Llama \cite{touvron2023llama} models, the baselines include three W4A4 quantization methods: SmoothQuant \cite{xiao2023SmoothQuant}, OmniQuant \cite{shao2024omniquant}, and Atom \cite{MLSYS2024_5edb57c0}. For OPT \cite{zhang2022opt} models, the baselines consist of OneBit \cite{xu2024onebit}, LRQuant \cite{zhao2024lrquant}, and GPTQ \cite{frantar2022gptq}. Performance is evaluated using Perplexity and Zero-Shot accuracy, while energy efficiency is measured in picojoules (pJ).

\subsection{Comparative study on Accuracy}

\begin{table*}[t]
  \caption{Zero-shot (\%) ($\uparrow$) performance of Llama2 and OPT models on six common sense tasks.}
  \label{tab:quant_performance_full}
  \centering
  \newcolumntype{C}[1]{>{\centering\arraybackslash}p{#1}}
  
  \begin{tabular}{ll *{7}{C{1.1cm}}}
    \toprule
    Model & Method & PIQA & ARC\_e & ARC\_c & BoolQ & HellaSwag & Winogrande & Avg. \\
    \midrule
    
    \multirow{8}{*}{Opt-1.3B} 
    & Fp16 & 72.42 & 50.80 & 29.69 & 57.68 & 53.73 & 59.67 & 54.00 \\
    \cmidrule(l){2-9} 
    & SpikeQuant & 71.49 & \textbf{49.66} & 27.22 & 53.27 & 51.46 & \textbf{58.56} & 51.94 \\
    & Quantization & 67.79 & 42.93 & 25.26 & 49.51 & 45.94 & 54.93 & 47.73 \\
    & OneBit & 62.57 & 41.25 & 24.06 & \textbf{59.45} & 34.26 & 51.14 & 45.46 \\
    & LRQuant & 60.23 & 43.22 & 19.80 & 57.98 & 30.71 & 51.78 & 43.95 \\
    & OPTQ & 68.34 & 46.17 & 27.65 & 51.57 & 46.53 & 55.26 & 49.25 \\
    \rowcolor{gray!10}
    & Kirin & \textbf{71.86} & 48.78 & \textbf{28.67} & 53.66 & \textbf{52.25} & 57.77 & \textbf{52.17} \\
    \midrule
    
    \multirow{8}{*}{Opt-2.7B} 
    & Fp16 & 74.81 & 54.34 & 31.31 & 60.28 & 60.59 & 60.93 & 57.04 \\
    \cmidrule(l){2-9}
    & SpikeQuant & 74.21 & 53.79 & \textbf{31.66} & 61.19 & \textbf{59.11} & 59.75 & 56.62 \\
    & Quantization & 71.16 & 50.84 & 28.75 & 59.85 & 55.57 & \textbf{60.46} & 54.44 \\
    & OneBit & 63.87 & 43.39 & 24.40 & 54.28 & 38.18 & 51.67 & 45.97 \\
    & LRQuant & 64.04 & 47.81 & 20.65 & 54.10 & 33.59 & 53.51 & 45.62 \\
    & OPTQ & 71.38 & 48.19 & 27.82 & 55.43 & 49.86 & 54.82 & 51.25 \\
    \rowcolor{gray!10}
    & Kirin & \textbf{74.81} & \textbf{53.91} & 31.48 & \textbf{61.69} & 58.83 & 59.91 & \textbf{56.77} \\
    \midrule
    
    \multirow{9}{*}{Llama2-7B} 
    & Fp16 & 77.37 & 52.53 & 41.38 & 73.12 & 72.99 & 66.85 & 64.04 \\
    \cmidrule(l){2-9}
    & SpikeQuant & \textbf{77.09} & \textbf{55.39} & 40.36 & 73.49 & 72.09 & 65.27 & 63.95 \\
    & Quantization & 75.46 & 53.96 & 40.19 & 68.07 & 69.21 & 61.56 & 61.41 \\
    & SmoothQuant & 63.11 & 40.03 & 31.57 & 58.47 & 43.38 & 52.80 & 48.23 \\
    & OmniQuant & 66.15 & 45.20 & 31.14 & 63.51 & 56.44 & 53.43 & 52.65 \\
    & Atom & 76.28 & 52.10 & 38.99 & 69.79 & 69.81 & 63.69 & 61.78 \\
    & SpikeLLM & 64.47 & 48.74 & 27.30 & 63.27 & 43.29 & 56.83 & 50.65 \\
    \rowcolor{gray!10}
    & Kirin & 76.77 & 54.55 & \textbf{40.53} & \textbf{73.91} & \textbf{72.10} & \textbf{66.22} & \textbf{64.01} \\
    \midrule
    
    \multirow{9}{*}{Llama2-13B} 
    & Fp16 & 79.05 & 59.85 & 44.62 & 68.53 & 76.22 & 70.09 & 66.39 \\
    \cmidrule(l){2-9}
    & SpikeQuant & 79.33 & 56.48 & \textbf{44.03} & 66.15 & 75.53 & 68.11 & 64.94 \\
    & Quantization & 76.71 & 55.13 & 41.47 & 67.03 & 73.09 & 65.19 & 63.10 \\
    & SmoothQuant & 64.47 & 41.75 & 30.89 & 62.29 & 46.68 & 51.70 & 49.63 \\
    & OmniQuant & 69.69 & 47.39 & 33.10 & 62.84 & 58.96 & 55.80 & 54.63 \\
    & Atom & 77.69 & \textbf{57.58} & 42.92 & \textbf{67.46} & 73.77 & 68.51 & 64.66 \\
    & SpikeLLM & 66.49 & 55.30 & 30.12 & 64.16 & 47.43 & 51.54 & 52.51 \\
    \rowcolor{gray!10}
    & Kirin & \textbf{79.35} & 56.40 & 43.94 & 66.21 & \textbf{75.94} & \textbf{68.98} & \textbf{65.14} \\
    
    \bottomrule
  \end{tabular}
\end{table*}

For performance evaluation, we report perplexity on WikiText-2 \cite{merity2017pointer} and C4 \cite{raffel2020exploring}, as well as zero-shot performance on PIQA \cite{bisk2020piqa}, ARC-easy/challenge \cite{clark2018think}, BoolQ \cite{clark2019boolq}, HellaSwag \cite{clark2018think}, and Winogrande \cite{sakaguchi2021winogrande}. Due to the limited number of SNN-based LLMs, we benchmark against a broad range of ANN quantization methods. This comparison is critical: since converting ANNs to SNNs inherently risks performance degradation, surpassing these quantization baselines serves as strong evidence that our proposed method achieves a lossless conversion.

As shown in Table \ref{tab:quant_performance_full}, our proposed Kirin achieves state-of-the-art performance across both Llama2 and OPT model families. In Table 1(a), for Llama2-7B on the Wiki dataset, Kirin achieves a perplexity of 5.71, which is remarkably close to the FP16 baseline (5.68) and significantly outperforms existing W4A4 quantization methods like SmoothQuant (22.62) and OmniQuant (14.61). Notably, compared to the advanced SNN baseline SpikeQuant (T=32), Kirin achieves superior accuracy with only half the inference latency (T=16). Table 1(b) further demonstrates the robustness of our method on OPT models. Kirin consistently outperforms aggressive low-bit quantization methods (e.g., OneBit, LRQuant) and surpasses SpikeQuant. The superior performance can be attributed to our comprehensive handling of outliers in both weights and activations, as indicated in the method configuration columns.

Table \ref{tab:quant_performance_full} reports the zero-shot accuracy on six common sense tasks. Kirin achieves near-lossless performance compared to the FP16 baseline, particularly on Llama2-7B (64.01\% vs. 64.04\%). Furthermore, it consistently outperforms both ANN quantization methods (e.g., SmoothQuant, OmniQuant) and state-of-the-art SNN baselines (e.g., SpikeQuant, Atom) across all model scales, demonstrating superior preservation of reasoning capabilities. Although SpikeQuant slightly outperforms Kirin on specific tasks (e.g., ARC\_e), this is attributed to SpikeQuant retaining floating-point precision for the attention matrix multiplication, whereas Kirin converts all matrix multiplications to spikes. Despite the challenge of full-network conversion, Kirin achieves a higher average accuracy. This success stems from our Spike Matrix Hybridization strategy, which selectively quantizes outliers to 8-bit precision. This approach effectively mitigates information loss in sensitive layers while maintaining a low time window, resulting in superior overall robustness.

\subsection{Comparative study on Energy}

Following prior works on SNN energy estimation \cite{yan2024reconsidering, wang2025energy, yan2025otters}, the total energy consumption of a linear attention module during a single inference pass is primarily composed of computation costs and data movement costs. The computation energy accounts for the MAC and ACC operations. The data movement energy covers the costs associated with reading weights from SRAM and moving data from the previous layer's output to the current layer for computation. Consequently, the total energy cost is calculated as the sum of the operation counts (MAC or ACC) multiplied by their respective energy costs per operation, plus the total number of data bits read or moved multiplied by the corresponding energy consumption per bit:
\begin{equation}
    E_{total} = \sum_{op \in \{MAC, ACC\}} N_{op} \cdot E_{op} + \sum_{d \in \{read, move\}} N_{bits}^d \cdot E_d
\label{eq17}
\end{equation}

\begin{figure}[t] 
    \centering
    \begin{subfigure}[b]{1.0\linewidth} 
        \centering
        \includegraphics[width=1.0\linewidth]{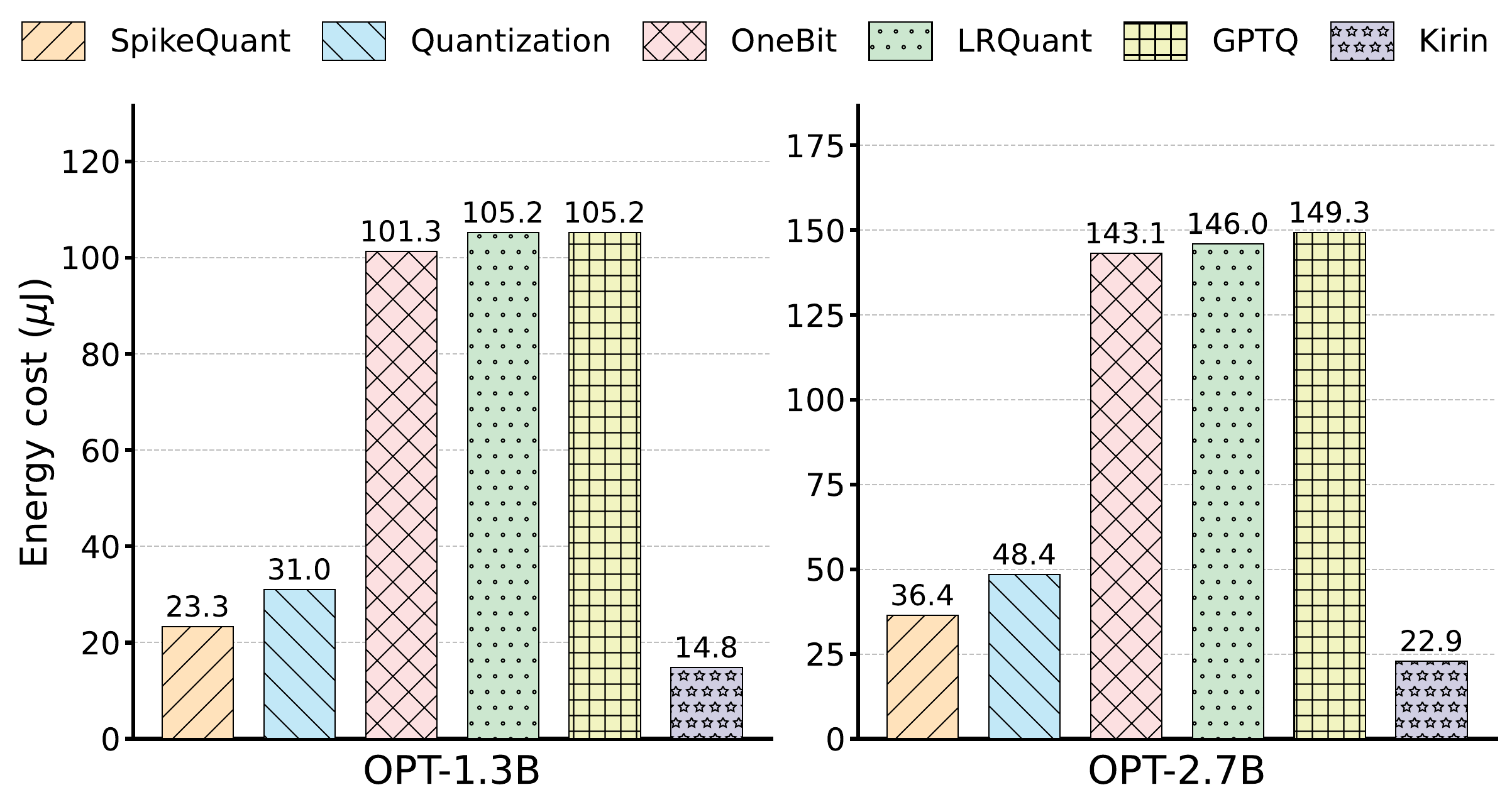}
        \caption{Energy cost comparison on OPT models.}
        \label{fig:opt_energy}
    \end{subfigure}
    
    \vspace{10pt} 
    
    \begin{subfigure}[b]{1.0\linewidth}
        \centering
        \includegraphics[width=1.0\linewidth]{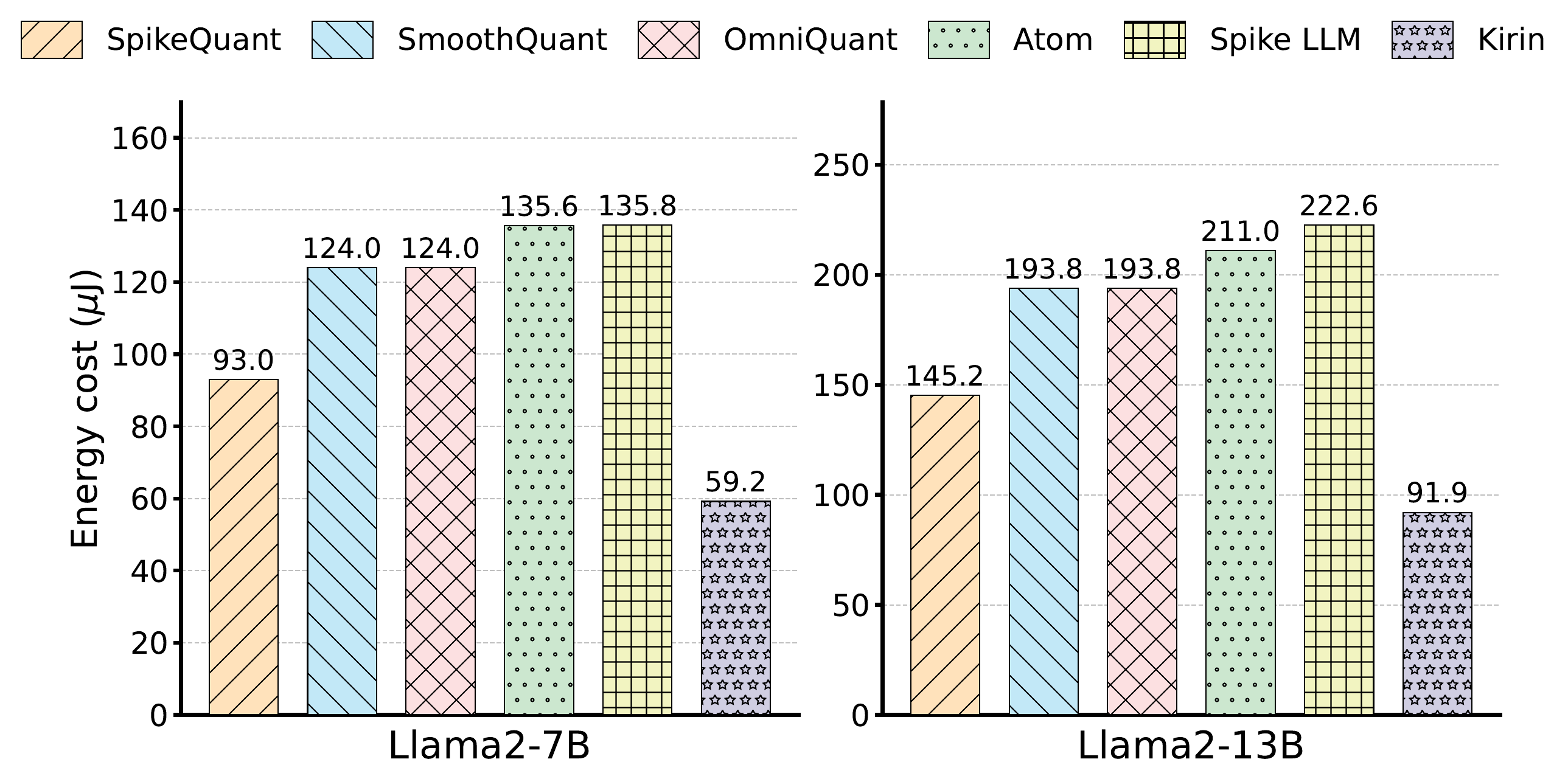}
        \caption{Energy cost comparison on Llama2 models.}
        \label{fig:llama_energy}
    \end{subfigure}
    
    \caption{Energy consumption analysis across different model architectures.}
    \label{fig:energy_comparison_all}
\end{figure}

Figure \ref{fig:energy_comparison_all} illustrates the energy consumption comparison between the proposed Kirin and other SNN or quantization methods within the linear attention block (four linear transformation and two attention transformation as shown in Figure \ref{fig:workflow}). Experimental results demonstrate that although a portion of INT values is retained within the spike matrix, the energy consumption increases by a mere 0.283 $\mu$J compared to a pure spike matrix. However, leveraged by the Spike Matrix Hybridization mechanism, the attention matrix multiplication can be flexibly converted into the SNN computational paradigm, thereby saving 17.458 $\mu$J in a single matrix multiplication operation. Consequently, Kirin achieves the lowest energy consumption among all compared algorithms, delivering energy savings of up to 84\%. Detailed energy calculation formulas and computation details are provided in Appendices A, B, and C.

\subsection{Ablation study}
Table \ref{tab:perplexity_ablation} illustrates that quantization is the primary source of error in ANN-to-SNN conversion. Naively quantizing all components degrades WikiText-2 perplexity from 5.68 to 6.844. While outlier-aware mixed quantization can mitigate this, simultaneously optimizing for static weights and dynamic activations remains challenging due to their conflicting distributions and high computational overhead.

However, our proposed Spike Matrix Hybridization overcomes this by unifying all operations—whether linear transformations (weight $\times$ activation) or attention mechanisms ($Q \times K$)—into a generic matrix multiplication framework. This perspective allows us to apply comprehensive outlier-aware optimization across all layers. As shown in the table, sequentially enabling outlier awareness for linear activations, weights, and attention matrices significantly recovers performance, reducing perplexity from 6.844 back to 5.798.

Finally, the introduction of the silence threshold ensures a seamless transition to the spiking domain. As evidenced by the final row, the performance change is exactly +0.000, confirming that our specific SNN conversion step is mathematically lossless.

\begin{table}[t]
  \caption{Ablation study on LLaMA-7B. Values in parentheses indicate the change relative to the previous row.}
  \label{tab:perplexity_ablation}
  \centering
  \renewcommand{\arraystretch}{1.2}
  
  \begin{tabular}{lcc}
    \toprule
    \multirow{2}{*}{Method} & \multicolumn{2}{c}{Perplexity ($\downarrow$)} \\
    \cmidrule(lr){2-3}
    & Wikitext & C4 \\
    \midrule
    
    FP16 & 5.68 & 7.08 \\
    \midrule
    
    +Linear projections quantization & 6.054 \footnotesize{(+0.374)} & 7.633 \footnotesize{(+0.553)} \\
    +Attention MatMul quantization   & 6.494 \footnotesize{(+0.440)} & 8.126 \footnotesize{(+0.493)} \\
    +MLP quantization                & 6.844 \footnotesize{(+0.350)} & 8.516 \footnotesize{(+0.390)} \\
    +Linear Activation outlier aware & 6.467 \footnotesize{(-0.377)} & 8.102 \footnotesize{(-0.414)} \\
    +Weight outlier aware            & 6.193 \footnotesize{(-0.274)} & 7.786 \footnotesize{(-0.316)} \\
    +Attention MatMul outlier aware  & 5.798 \footnotesize{(-0.395)} & 7.324 \footnotesize{(-0.462)} \\
    +SNN conversion (Kirin)                  & 5.798 \footnotesize{(+0.000)} & 7.324 \footnotesize{(+0.000)} \\
    
    \bottomrule
  \end{tabular}
\end{table}

\section{Related Works on ANN-to-SNN Conversion}

Recently, brain-inspired SNNs \cite{gerstner2014neuronal} have emerged to bridge the energy gap between ANN inference and human cognition, and are increasingly recognized as a promising next-generation artificial intelligence paradigm due to their energy-efficient inference capabilities. Consequently, ANN-to-SNN conversion techniques have garnered escalating research interest, yielding substantial advancements in the field.

With the rapid advancement of machine learning, adapting the ubiquitous Transformer architecture from ANNs to SNNs has emerged as a prominent research hotspot \cite{10388996,10.1016/j.neucom.2024.127279,wang2025spiking}. Spikformer \cite{zhou2023spikformer} eliminates the Softmax mechanism to linearize attention, allowing matrix multiplication reordering. By computing $K \times V$ first, it converts heavy computations into energy-efficient bitwise ANDs and accumulations. Building upon this foundation, subsequent studies have further optimized the spiking integrity of the architecture. Addressing the limitation in Spikformer where non-spiking computations were retained in residual connections and ConvBN layers,  Spike-driven Transformer \cite{NEURIPS2023_ca0f5358} restructured the residual connections to ensure binary spike signal propagation throughout the entire network. Additionally, Guo et al. \cite{Guo_2025_CVPR} introduced a mixed-precision approach, retaining $K$ in floating-point format while quantizing $V$ to ternary values (-1, 0, 1) to balance accuracy and efficiency. However, a critical limitation persists across these existing works: the linear projection layers within the Transformer (responsible for generating $Q$, $K$, and $V$) remain unprocessed by SNN mechanisms. These layers continue to rely on energy-intensive floating-point matrix multiplications, leaving significant room for further energy reduction in SNN-based Transformers.

Traditional ANN-to-SNN methods, limited to smaller architectures like ResNet, are inapplicable to massive LLMs. Consequently, recent works like SpikeLLM \cite{xing2024spikellm} and SpikeQuant \cite{wang2025energy} adopt a quantization-based conversion paradigm. While SpikeQuant ensures mathematical equivalence by deriving firing thresholds from quantization scales, both methods overlook the long inference windows caused by high bit-widths. Furthermore, their reliance on rate encoding leads to excessive spike rates, significantly increasing computational and energy overheads.

\section{Conclusion}
In this paper we present Kirin, the first framework to achieve accuracy-lossless ANN-to-SNN conversion through integer-spike hybridization. By pioneering a hybrid execution strategy, Kirin decomposes matrix multiplications into a hybrid path where dominant spike-train accumulations are complemented by a minimal fraction of integer multiplications for outliers. To support this, we introduce Spike Matrix Hybridization to minimize integer overhead and a Silence Threshold mechanism to ensure mathematical equivalence to the original ANN. This design effectively resolves the long time window bottleneck induced by high bit-width quantization. On OPT-2.7B, Kirin attains near-FP16 accuracy under W4A(4\&8) settings while drastically reducing the time window from 256 to 16. Consequently, energy consumption drops from 149.3 $\mu$J to 22.9 $\mu$J.


\bibliographystyle{unsrt}
\bibliography{main}

\begin{table*}[t]
\centering
\begin{tabular}{l|l} 
\hline
\textbf{Symbol} & \textbf{Description} \\
\hline
$B, H_{in}, H_{out}$ & Batch size, input hidden dimension, and output hidden dimension \\
$S$ & Sequence lengths of inputs corresponding to Query, Key, Value and Output \\
$b_w, b_a$ & Bitwidths of weights and activations \\
$E_{MAC}, E_{ACC}$ & Energy costs per Multiply-Accumulate (MAC) and Accumulate (ACC) operation \\
$E_{read}, E_{move}$ & Energy consumption associated with memory read and data movement \\
$\gamma$ & Number of outlier channels within a matrix group \\
$\beta$ & Number of outlier parameters in the spike matrix retained in INT format \\
$T^{high}, T^{low}$ & Spike time windows for outlier and normal parameters \\
$S_r^{high}, S_r^{low}$ & Spike rates corresponding to outlier and normal parameters \\
\hline
\end{tabular}
\caption{Summary of Notations}
\label{tab:notations}
\end{table*}


\newpage
\appendix

\section{Notation Definition}
For clarity and ease of reference regarding the mathematical formulations and energy estimations, Table \ref{tab:notations} provides a comprehensive summary of the notations used throughout this paper. The symbols are organized to cover three primary aspects: general model architecture dimensions (e.g., $B, H_{in}, S$), hardware-related metrics such as bitwidths and energy consumption constants for memory and computation ($E_{MAC}, E_{read}$), and the specific hyperparameters introduced by our framework. The latter includes parameters governing the outlier-aware mixed quantization ($\gamma, \beta$) as well as the configuration for the dual-window spiking mechanism ($T^{high/low}, S_r^{high/low}$).

\section{Energy Cost Formulation}
In this section, we present the detailed methodology for calculating energy costs. The fundamental principle involves multiplying the number of operations in a linear or attention matrix transformation by the corresponding unit energy consumption. In LLM architectures, such as Llama and OPT, the dimensions of the linear activation tensor are typically $[B, S, H_{in}]$, while the weight matrix dimensions are $[H_{in}, H_{out}]$. Following the linear transformation, the output tensor dimensions become $[B, S, H_{out}]$, which subsequently serves as the input for the attention mechanism. Given the disparities in tensor dimensions and computational characteristics between linear transformations and attention mechanisms, we derive separate formulas for the energy consumption of linear transformations ($E_{Linear}$) and attention mechanisms ($E_{Attn}$). Furthermore, in the context of outlier-aware mixed-precision quantization, the high-bit and low-bit components must be calculated independently due to significant differences in element counts and per-bit energy consumption; these are denoted as $E^{high}$ and $E^{low}$, respectively. Based on these computational paradigms, the energy consumption can be categorized into the following four classes:

\textbf{Baseline Quantization Energy:} In the standard quantization paradigm, the linear layer involves the multiplication of a weight matrix with dimensions $[H_{in}, H_{out}]$ and an activation tensor with dimensions $[B, S, H_{in}]$, resulting in a total of $B \cdot S \cdot H_{in} \cdot H_{out}$ Multiply-Accumulate (MAC) operations. Beyond computational energy, the total cost must account for memory access overhead, specifically the energy required to read weights from memory and transfer activations from the preceding layer. In contrast, matrix multiplication within the attention mechanism entails the product of two tensors, each with dimensions $[B, S, H_{out}]$, yielding $B \cdot S^2 \cdot H_{out}$ MAC operations. This step requires no weight memory access; rather, it only incurs the energy expenditure associated with moving the two intermediate results generated by the previous linear transformation.

\begin{equation}
\begin{split}
        LE_q = &\underbrace{B\cdot S\cdot H_{in}\cdot  H_{out}\cdot E_{MAC}}_{E_{compute}} \\
    &+\underbrace{B\cdot S \cdot H_{in}\cdot H_{out}\cdot b_w \cdot E_{read}}_{E_{read\_data}} \\
    &+\underbrace{  B\cdot S\cdot H_{in}\cdot H_{out}\cdot b_a \cdot E_{move}}_{E_{move\_data}}
\end{split}
\end{equation}

\begin{equation}
\begin{split}
        AE_q = &\underbrace{B\cdot S^2\cdot H_{out}\cdot E_{MAC}}_{E_{compute}} 
    +\underbrace{ 2 \cdot B\cdot S\cdot H_{out}^2\cdot b_a \cdot E_{move}}_{E_{move\_data}}
\end{split}
\end{equation}

\textbf{Outlier-Aware Quantization Energy:} In the context of mixed-precision quantization, the computational framework remains largely consistent with the standard quantization paradigm. The primary distinction lies in the necessity to independently compute the energy consumption for high-bit and low-bit quantized components, respectively, while the remaining operational logic and overhead calculations are preserved.
\begin{equation}
    LE_{mq} = LE_{mq}^{high}+LE_{mq}^{low}
\end{equation}
\begin{equation}
\begin{split}
        LE_{mq}^{high} &= \underbrace{B\cdot S\cdot H_{out}\cdot \gamma \cdot E_{MAC}^{high}}_{E_{compute}} \\
    &+\underbrace{B\cdot S\cdot\gamma\cdot H_{out}\cdot b_w \cdot E_{read} }_{E_{read\_data}} \\
    &+  \underbrace{B\cdot S\cdot \gamma\cdot H_{out}\cdot b_a^{high} \cdot E_{move}}_{E_{move\_data}}
\end{split}
\end{equation}
\begin{equation}
\begin{split}
        LE_q^{low} = &\underbrace{B\cdot S\cdot H_{out}\cdot (H_{in}-\gamma) \cdot E_{MAC}^{low}}_{E_{compute}} \\
    &+\underbrace{B\cdot S\cdot(H_{in}-\gamma)\cdot H_{out}\cdot b_w \cdot E_{read}}_{E_{read\_data}} \\
    &+\underbrace{B\cdot S\cdot (H_{in}-\gamma)\cdot H_{out}\cdot b_a^{low} \cdot E_{move}}_{E_{move\_data}}
\end{split}
\end{equation}
\begin{equation}
    AE_{mq} = AE_{mq}^{high}+AE_{mq}^{low}
\end{equation}

\begin{equation}
\begin{split}
        AE_{mq}^{high} = &\underbrace{B\cdot S^2\cdot \gamma \cdot E_{MAC}^{high}}_{E_{compute}}
    + \underbrace{2\cdot B\cdot S\cdot \gamma\cdot H_{out}\cdot b_a^{high} \cdot E_{move}}_{E_{move\_data}}
\end{split}
\end{equation}
\begin{equation}
\begin{split}
        AE_{mq}^{low} = &\underbrace{B\cdot S^2\cdot (H_{out}-\gamma) \cdot E_{MAC}^{low}}_{E_{compute}} \\
        &+\underbrace{2\cdot B\cdot S\cdot (H_{out}-\gamma)\cdot H_{out}\cdot b_a^{low} \cdot E_{move}}_{E_{move\_data}}
\end{split}
\end{equation}

\textbf{SNN Inference Energy:} In SNNs based on mixed-precision quantization, it is essential to account for the additional energy expenditure associated with encoding quantized values into binary spike trains (e.g., Rate encoding or TTFS Encoding). Furthermore, given that data representation within SNNs is fundamentally binary, the energy assessments for computation and data movement must focus on discrete individual spikes rather than full multi-bit quantized values. Since the SNN baselines evaluated in this work do not convert the attention matrix multiplication into the spiking domain, the energy consumption for the attention component is equivalent to $AE_{mq}$.
\begin{equation}
    LE_s = LE_s^{high}+LE_s^{low}
\end{equation}

\begin{equation}
\begin{split}
    LE_s^{high} &=
    \underbrace{B\cdot S\cdot H_{out} \cdot \gamma \cdot T^{high} \cdot S_r^{high}\cdot \big (E_{ACC}^{high}+E_{MAC}^{high}\big)}_{E_{compute}} \\
    &+\underbrace{B\cdot S\cdot\gamma\cdot H_{out}\cdot T^{high} \cdot S_r^{high} \cdot b_w \cdot E_{read}}_{E_{data\_read}} \\
    &+\underbrace{B\cdot S\cdot \gamma \cdot H_{out} \cdot T^{high} \cdot S_r^{high} \cdot E_{move}}_{E_{data\_move}}
\end{split}
\end{equation}
\begin{equation}
\begin{split}
    LE_s^{low} &=\underbrace{B\cdot S\cdot  H_{out} \cdot (H_{in}-\gamma) \cdot T^{low} \cdot S_r^{low}\cdot \big (E_{ACC}^{low}+E_{MAC}^{low}\big)}_{E_{compute}} \\
    &+\underbrace{B\cdot S\cdot(H_{in}-\gamma)\cdot H_{out} \cdot T^{low} \cdot S_r^{low}\cdot b_w \cdot E_{read}}_{E_{data\_read}} \\
    &+\underbrace{ B\cdot S \cdot (H_{in}-\gamma)\cdot H_{out}\cdot T^{low} \cdot S_r^{low} \cdot E_{move}}_{E_{data\_move}}
\end{split}
\end{equation}



\begin{table*}[t]
  \caption{Energy Ratio of Different Computing Units and Data Movement Operations (Normalized to 4-4-4 ACC)}
  \label{tab:energy_computing_data_movement}
  \centering
  \newcolumntype{C}[1]{>{\centering\arraybackslash}p{#1}}
  \begin{tabular}{C{1.2cm} *{9}{C{1.1cm}} *{2}{C{1.1cm}}}
    \toprule
    & \multicolumn{9}{c}{Computing} & \multicolumn{2}{c}{Data movement} \\
    \cmidrule(lr){2-10}\cmidrule(lr){11-12}
    Process 
      & 4-4-4 ACC 
      & 5-5-5 ACC 
      & 1-4-16 MAC  
      & 4-4-32 MAC 
      & 4-5-32 MAC
      & 4-8-32 MAC 
      & 1-16-32 MAC 
      & 2-16-32 MAC 
      & 3-16-32 MAC 
      & Read/bit 
      & Move/bit \\
    \midrule
    Energy 
       & 1.00 
       & 1.18 
       & 4.06  
       & 8.66 
       & 9.24 
       & 10.94 
       & 10.89 
       & 11.46 
       & 13.28 
       & 6.04 
       & 11.04 \\
    \bottomrule
  \end{tabular}
\label{tab:energy_value}
\end{table*}

\begin{table*}[t]
    \centering
    \caption{Average ratio and number of outliers ($\gamma$) values and integer values ($\beta$) in OPT and Llama2 models}
    \label{outliers_number}
    \setlength{\tabcolsep}{6pt}
    \begin{tabular}{lrrrrrrrr}
        \toprule
        Methods &
        \multicolumn{2}{c}{OPT1.3B (H=2048)} &
        \multicolumn{2}{c}{OPT2.7B (H=2560)} &
        \multicolumn{2}{c}{Llama2-7B (H=4096)} &
        \multicolumn{2}{c}{Llama2-13B (H=5120)} \\
        \cmidrule(lr){2-3} \cmidrule(lr){4-5} \cmidrule(lr){6-7} \cmidrule(lr){8-9}
        & Ratio & Number
        & Ratio & Number
        & Ratio & Number
        & Ratio & Number \\
        \midrule
        SpikeQuant & 2.45\% & $\gamma$=50  & 2.15\% & $\gamma$=55  & 2.68\% & $\gamma$=110 & 2.30\% & $\gamma$=118 \\
        Atom       & 6.25\% & $\gamma$=128 & 5.00\% & $\gamma$=128 & 3.12\% & $\gamma$=128 & 2.50\% & $\gamma$=128 \\
        SpikeLLM   & 10.00\% & $\gamma$=205 & 10.00\% & $\gamma$=256 & 10.00\% & $\gamma$=410 & 10.00\% & $\gamma$=512 \\
        Kirin  & 1.96\% & $\beta$=40 & 1.61\% & $\beta$=41 & 2.14\% & $\beta$=87 & 1.95\% & $\beta$=99 \\
        \bottomrule
    \end{tabular}
\label{tab:salient_ratio}
\end{table*}

\textbf{Kirin Energy:} The proposed Kirin architecture strategically retains quantized values that would typically require long time windows in Integer  format. Consequently, this approach eliminates the spike-based energy consumption associated with long time windows ($LE_s^{high}$ and $AE_s^{high}$). Instead, it introduces a marginal energy cost attributed to integer multiplication, denoted as $LE_{mq}$ and $AE_{mq}$. Furthermore, within the calculations for $LE_{mq}^{high}$ and $AE_{mq}^{high}$, the parameter $\gamma$, which originally represented the count of outliers, is substituted by $\beta$, representing the count of preserved integers. Similarly, for $LE_{s}^{low}$ and $AE_{s}^{low}$, the terms $(H_{in}-\gamma)$, representing the count of low-bit normal values, is replaced by $(H_{in}-\beta)$.
\begin{equation}
\begin{split}
    LE_{SiM} &= LE_{mq}^{high}+LE_s^{low} \\
    & \underbrace{B\cdot S\cdot H_{out}\cdot \beta \cdot E_{MAC}^{high}}_{E_{compute}} \\
    &+\underbrace{B\cdot S\cdot \beta\cdot H_{out}\cdot b_w \cdot E_{read} }_{E_{read\_data}} +  \underbrace{B\cdot S\cdot \beta \cdot H_{out}\cdot b_a^{high} \cdot E_{move}}_{E_{move\_data}} \\
    &+\underbrace{B\cdot S\cdot  H_{out} \cdot (H_{in}-\beta) \cdot T^{low} \cdot S_r^{low}\cdot \big (E_{ACC}^{low}+E_{MAC}^{low}\big)}_{E_{compute}} \\
    &+\underbrace{B\cdot S\cdot(H_{in}-\beta)\cdot H_{out}\cdot T^{low} \cdot S_r^{low} \cdot b_w \cdot E_{read}}_{E_{data\_read}} \\
    &+\underbrace{ B\cdot S \cdot (H_{in}-\beta)\cdot H_{out} \cdot T^{low} \cdot S_r^{low} \cdot E_{move}}_{E_{data\_move}}
\end{split}
\end{equation}
\begin{equation}
\begin{split}
        AE_{SiM} &= AE_{mq}^{high}+AE_s^{low} \\
    &\underbrace{B\cdot S^2\cdot \beta \cdot E_{MAC}^{high}}_{E_{compute}}
    + \underbrace{2\cdot B\cdot S\cdot H_{out}\cdot \beta\cdot b_a^{high} \cdot E_{move}}_{E_{move\_data}} \\
    &+\underbrace{B\cdot S^2\cdot  (H_{out}-\beta) \cdot T^{low} \cdot S_r^{low}\cdot \big (E_{ACC}^{low}+E_{MAC}^{low}\big)}_{E_{compute}} \\
    &+\underbrace{ 2\cdot B\cdot S \cdot (H_{out}-\beta) \cdot H_{out}\cdot T^{low} \cdot S_r^{low} \cdot E_{move}}_{E_{data\_move}}
\end{split}
\end{equation}

\section{Energy Cost Calculation Details}

To fairly evaluate the energy efficiency of the proposed Kirin against other quantization and SNN baselines, we conducted a detailed estimation based on the formulation derived in Appendix A and B. The calculation relies on specific hardware (22nm commercial chip) energy ratios listed in Table \ref{tab:energy_computing_data_movement}. All energy values are normalized to the energy cost of a 4-bit addition (4-4-4 ACC). As SNN-based operations primarily utilize accumulators, the 4-bit ACC serves as our baseline unit ($1.00\times$), whereas standard quantization methods rely on Multiply-Accumulate (MAC) operations, which consume significantly more energy (e.g., $8.66\times$ for a 4-4-32 MAC). Additionally, we account for the energy cost of reading weights and moving data, which are critical factors in large language models.

\begin{figure*}[t]
    \centering
    
    
    \begin{subfigure}[b]{0.48\textwidth}
        \centering
        \includegraphics[width=\linewidth]{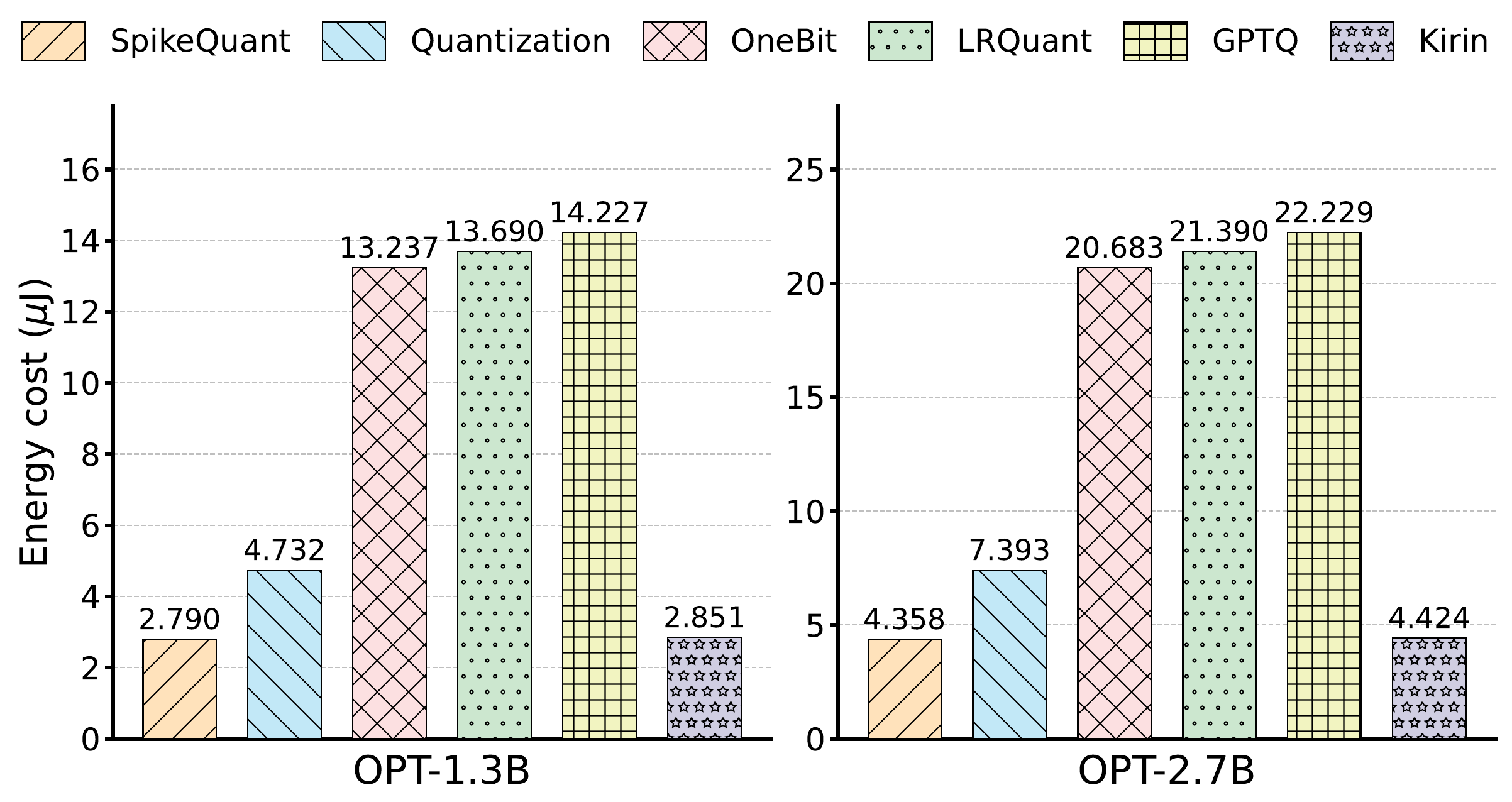}
        \caption{OPT Energy (Linear)}
        \label{fig:opt_linear}
    \end{subfigure}
    \hfill 
    \begin{subfigure}[b]{0.48\textwidth}
        \centering
        \includegraphics[width=\linewidth]{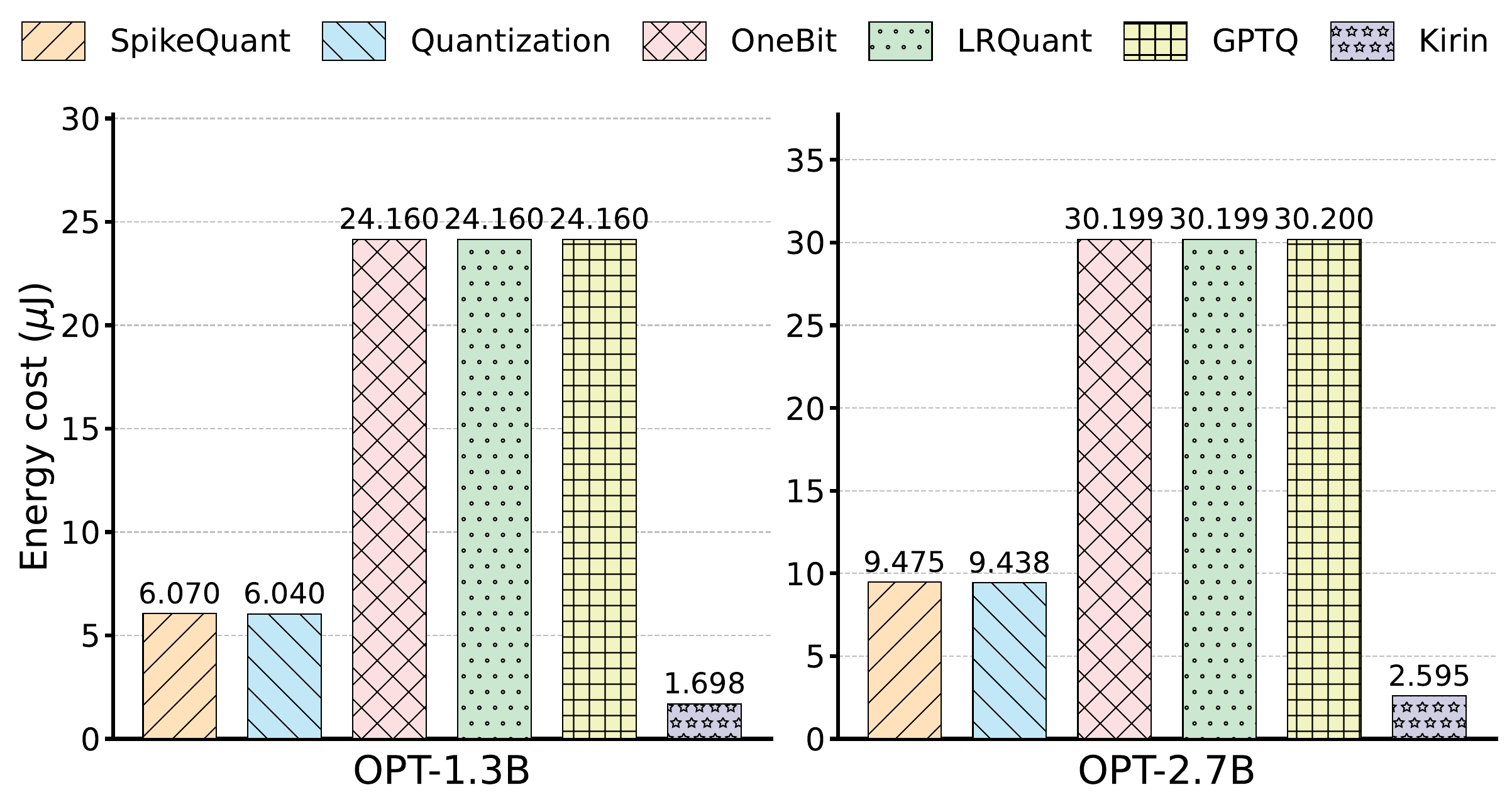}
        \caption{OPT Energy (Attention)}
        \label{fig:opt_attn}
    \end{subfigure}
    
    \vspace{1em} 
    
    
    \begin{subfigure}[b]{0.48\textwidth}
        \centering
        \includegraphics[width=\linewidth]{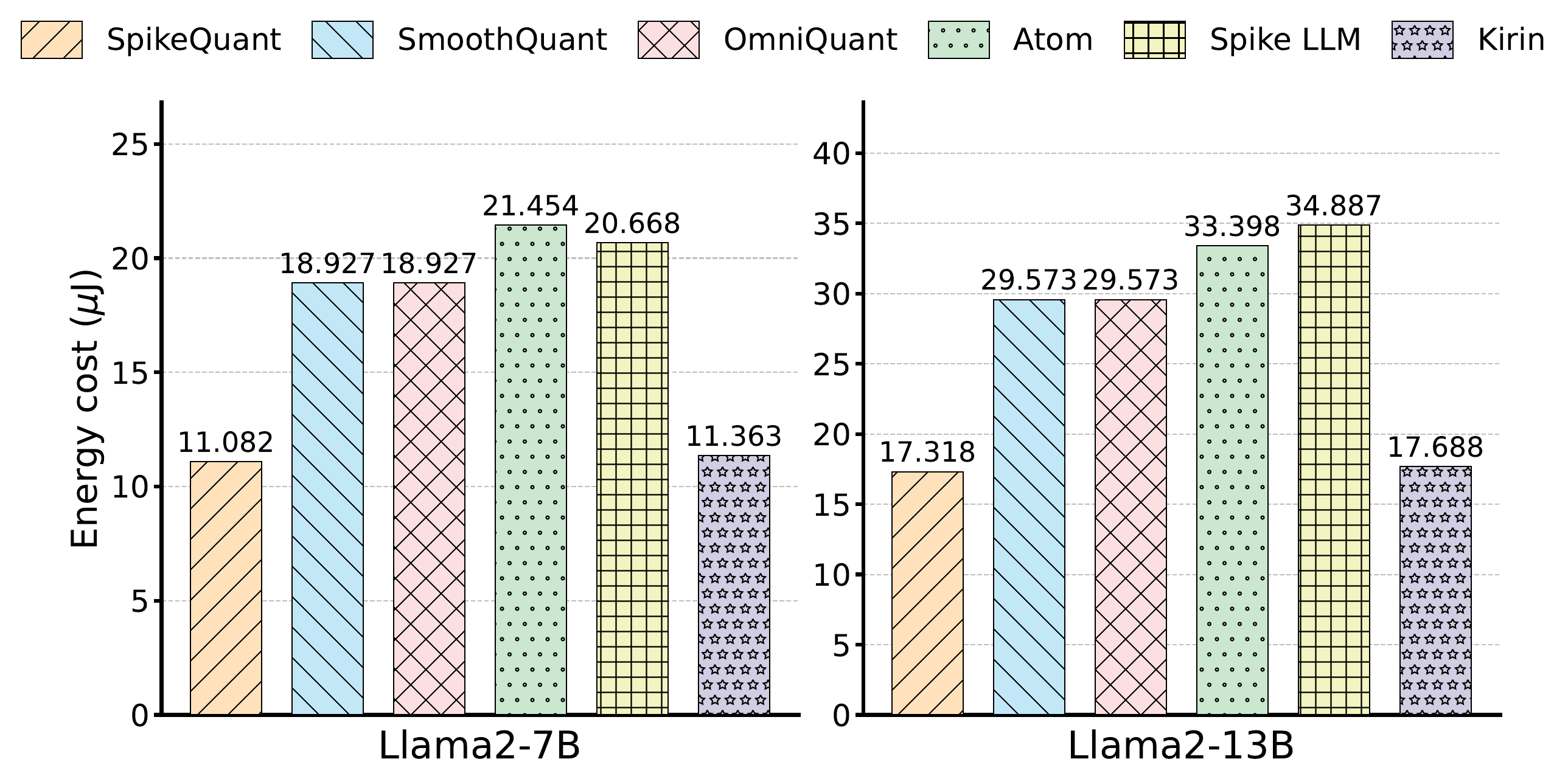}
        \caption{Llama Energy (Linear)}
        \label{fig:llama_linear}
    \end{subfigure}
    \hfill 
    \begin{subfigure}[b]{0.48\textwidth}
        \centering
        \includegraphics[width=\linewidth]{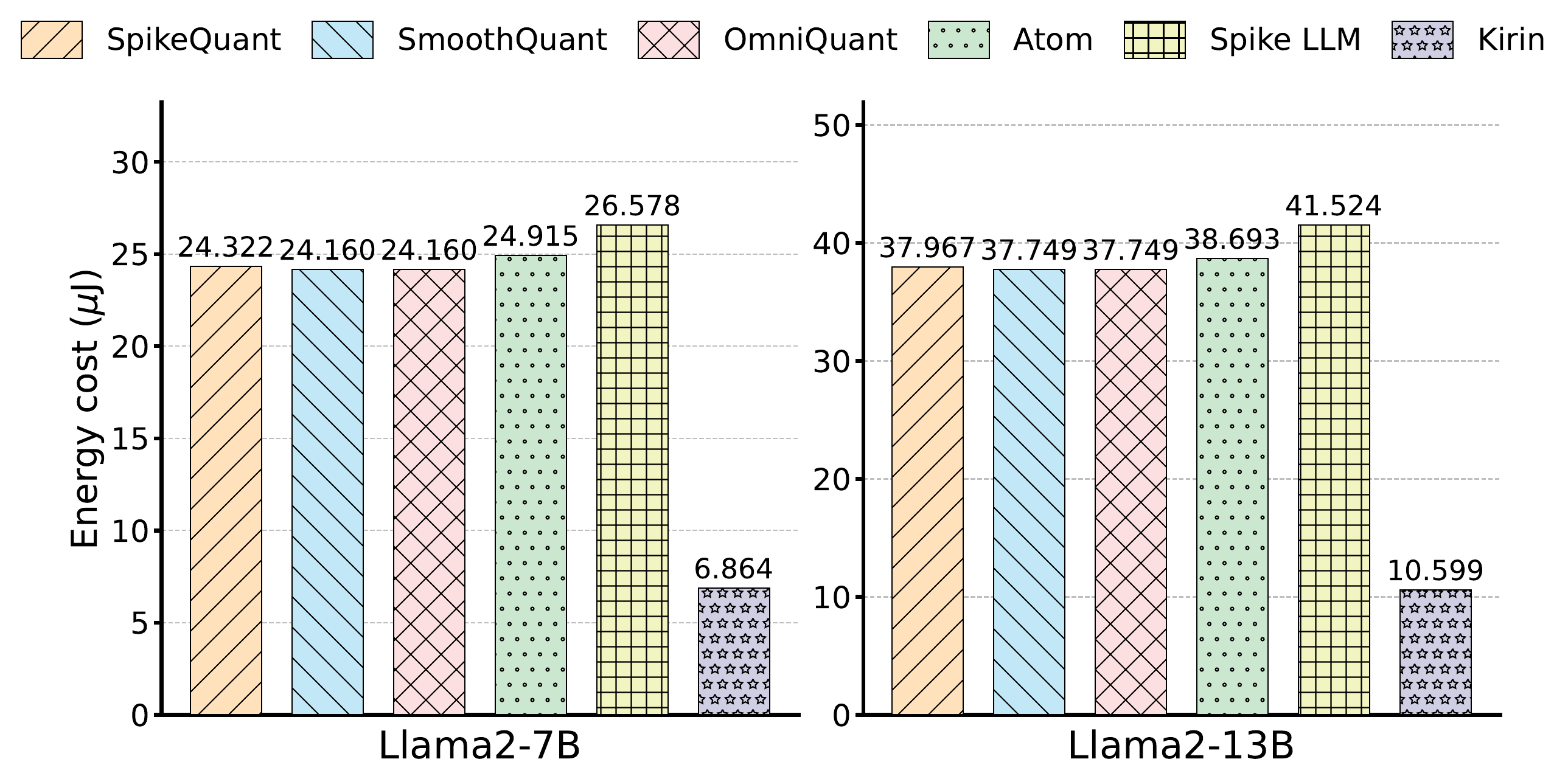}
        \caption{Llama Energy (Attention)}
        \label{fig:llama_attn}
    \end{subfigure}

    \caption{Comparison of energy consumption. Top row: OPT models; Bottom row: Llama models.}
    \label{fig:four_images}
\end{figure*}

The energy cost is strictly dependent on the sparsity and outlier distribution within the specific LLMs. Table \ref{outliers_number} summarizes the average outlier ratios and counts for the OPT and Llama2 model families used in our calculations. Parameters such as the outlier count ($\gamma$) dictate the number of neurons that must be handled by the high-precision branch in outlier-aware frameworks. The specific firing rates and integer values ($\beta$) presented in Table \ref{outliers_number} are plugged into the energy equations to determine the weighted sum of ACC and MAC operations.

A critical distinction in our evaluation concerns the handling of the Attention matrix transformation. Since the SNN baselines evaluated in this work do not convert the attention matrix multiplication into the spiking domain, the energy consumption for the attention component is equivalent to standard matrix multiplication ($AE_{mq}$). Consequently, the energy cost for SNN baselines in Attention layers remains high, comparable to standard quantization methods, as they cannot leverage the efficiency of sparse additions in this specific module.

Figure \ref{fig:four_images} visualizes the final calculated energy costs. In the Linear layers (Figs. \ref{fig:opt_linear} and \ref{fig:llama_linear}), the results demonstrate that the proposed integer-spike hybrid mode incurs negligible energy overhead. As shown, Kirin’s energy consumption remains comparable to the pure SNN baseline, SpikeQuant, while still being significantly lower than the 13–14 $\mu J$ range of 4-bit quantization methods (e.g., 2.85 $\mu J$ for Kirin on OPT-1.3B). However, the strategic value of this hybrid design is fully realized in the Attention layers (Figs. \ref{fig:opt_attn} and \ref{fig:llama_attn}). By explicitly handling outliers with integer arithmetic, Kirin enables the vast majority of attention parameters to be converted into energy-efficient spiking operations, whereas baselines are often forced to revert to energy-intensive standard MACs. Consequently, Kirin consumes only $\approx$ 6.86 $\mu J$ in Llama2-7B attention operations, compared to over 24 $\mu J$ for other baselines. Crucially, this hybrid approach facilitates a drastic reduction in the time window ($T$) by resolving the outlier bottleneck. Overall, Kirin achieves an optimal trade-off: it preserves the intrinsic energy efficiency of SNNs while significantly lowering latency and unlocking high-performance conversion for complex attention mechanisms.

\end{document}